%% file: paper.tex
\documentclass[journal]{IEEEtran}
\usepackage{amsmath,amsfonts,nicefrac}
\usepackage{algorithmic}
\usepackage{algorithm}
\usepackage{array}
\usepackage[caption=false,font=normalsize,labelfont=sf,textfont=sf]{subfig}
\usepackage{textcomp}
\usepackage{stfloats}
\usepackage{url}
\usepackage{verbatim}
\usepackage{graphicx}
\usepackage{cite}

\usepackage{orcidlink}

\begin{document}
\title{Transformers Meet Relational Databases}

\author{Jakub Peleška\textsuperscript{\orcidlink{0009-0000-8561-8106}}, Gustav Šír\textsuperscript{\orcidlink{0000-0001-6964-4232}}  
\thanks{Jakub Peleška and Gustav Šír are with the Department of Computer Science, Czech Technical University in Prague, 12135 Prague, Czechia (email: jakub.peleska@fel.cvut.cz, gustav.sir@cvut.cz)}}

\markboth{Transformers Meet Relational Databases - A Preprint}{}

\maketitle

\begin{abstract}
Transformer models have continuously expanded into all machine learning domains convertible to the underlying sequence-to-sequence representation, including tabular data. However, while ubiquitous, this representation restricts their extension to the more general case of \textit{relational databases}. In this paper, we introduce a modular neural message-passing scheme that closely adheres to the formal relational model, enabling \textit{direct} end-to-end learning of tabular Transformers from database storage systems. We address the challenges of appropriate learning data representation and loading, which are critical in the database setting, and compare our approach against a number of representative models from various related fields across a significantly wide range of datasets. Our results demonstrate a superior performance of this newly proposed class of neural architectures.
\end{abstract}

% Note that keywords are not normally used for peer-review papers.
% \begin{IEEEkeywords}
% relational databases, transformers, relational learning, message passing, graph neural networks (GNN)
% \end{IEEEkeywords}

\IEEEpeerreviewmaketitle

\section{Introduction}
% The very first letter is a 2 line initial drop letter followed by the rest of the first word in caps.
\IEEEPARstart{W}{hile} the approaches to mathematical modeling of complex systems, ranging from control theory to machine learning (ML), evolved in various independent ways, one aspect remained almost universal — the \textit{data representation}. Irrespective of the used models, from decision trees to neural networks, virtually all ML libraries expect input samples in the form of fixed-size numeric tensors, most often just (feature) vectors. Assuming the data samples as independent points in $n$-dimensional spaces is extremely convenient and allows for building directly upon the elegant foundations of linear algebra and multivariate statistics~\cite{james2013introduction}. However, actual real-world data is not stored in numeric vectors or tensors but mostly in the interlinked structures of internet pages, knowledge graphs, and, particularly, \textit{relational databases}. Indeed, while there are numerous data storage formats, the traditional relational database management systems (RDBMS) arguably dominate the industry, from medicine and engineering to enterprise application domains~\cite{halpin2010information}.

In recent years, we have witnessed \textit{deep learning} to quickly dominate all perceptual domains, from vision and speech to language. Nevertheless, it remains very rare to encounter neural models on the classic \textit{tabular data} with heterogeneous features, where standard statistical models, mainly various decision tree ensembles~\cite{friedman2001greedy}, still appear to lead the benchmarks~\cite{shwartz2022tabular}. Improving the performance of the neural models, primarily the omnipresent \textit{Transformer} architecture~\cite{vaswani2017attention}, on tabular datasets gains increasing amounts of attention, sometimes quoted as the ``last unconquered castle'' for deep learning~\cite{kadra2021well}. Nevertheless, generalizing Transformers from the tabular to the full \textit{relational} data model posits arguably an even bigger challenge.

In this paper, we introduce a new class of such deep learning architectures aimed directly at relational database representation while utilizing insights from the established field of \textit{relational learning}~\cite{Getoor2007}, which is concerned with such generalizations of statistical models. 

The core contribution of our work, put into context of related work in Sec.~\ref{sec:constributions}, is the design of a new neural message-passing scheme following the formal relational model while deeply integrating the existing (tabular) Transformer architectures. The implementation of the proposed framework is readily available at Github.~\footnote{\url{https://github.com/jakubpeleska/deep-db-learning}}

\section{Related Work}
\label{sec:related}

While the body of work on using deep learning with relational databases themselves is extremely scarce, there are established machine learning areas that either use neural models on simpler data structures or address relational structures with other (non-neural) models. In this section, we first briefly review these fields, often overlooked in deep learning, to properly position the contribution of our work (Sec.~\ref{sec:constributions}).

\subsection{Tabular models}

Tabular neural models~\cite{badaro2023transformers} are concerned with transferring deep learning strategies into the (classic) tabular data setting, currently still largely dominated by standard statistical models, such as gradient-boosted trees~\cite{friedman2001greedy}. These commonly aim to amend the Transformer architecture~\cite{vaswani2017attention} to better fit the complex, often heterogeneous and discrete, attribute structure of the tabular data.
Some notable models in this category include the TabNet~\cite{arik2020tabnet}, which uses a custom-modified transformer-based architecture; TabTransfomer~\cite{Huang2020}, which focuses on categorical values while utilizing the original Transformer Encoder structure; SAINT~\cite{somepalli2021saint}, which introduced the concept of inter-sample attention; and Trompt~\cite{chen2023trompt}, which takes inspiration from prompt learning of language models.
We note that these tabular Transformers are sometimes (confusingly) referred to as ``relational.'' However, they do not follow the actual relational (database) model and cannot be (directly) used as such.

\subsection{Statistical relational learning}

For decades~\cite{cropper2020turning}, proper learning with actual relational representations has been the concern of the little-known field of Relational machine learning~\cite{de2008logical}. It builds heavily on the formalism of first-order logic (FOL)~\cite{gallier2015logic}, in which the tabular representation and the corresponding models are effectively viewed as {\textit{propositional}}, while the database representation, corresponding formally to a subset of FOL, requires \textit{relational} generalization(s) of such models.
Many such FOL-based methods have been proposed, mostly following the paradigm of Inductive Logic Programming (ILP)~\cite{muggleton1994inductive}, later extended with probabilistic methods under the umbrella of Statistical Relational Learning (SRL)~\cite{Getoor2007}. The most appropriate SRL works capable of learning from database representations then follow the paradigm of ``lifting,''~\cite{LIFTED} referring to the generalization of classic statistical models into the relational setting.
However, building on the FOL foundations, the SRL models typically do not scale well and, importantly, do not offer the latent representation learning capabilities of neural networks.

\subsection{Propositionalization}

From the SRL view, the Tabular Transformers address the exact same representation expressiveness as their classic tree-based counterparts they aim to surpass. The tabular, also known as ``attribute-value,'' data format 
% with columns corresponding to data features and rows to data samples, 
is an established ML representation perpetuating the whole field.
% , expected at input in virtually any ML library. 
% However, most real-world data are not stored as feature vectors but in various interlinked structures, most notably relational databases.
While much of the real-world data structures, such as relational databases, do \textit{not} fit into this representation, a natural urge arises to transform such structures into the expected format and proceed with the standard models. This practice, generally referred to as \textit{propositionalization}~\cite{kramer2001propositionalization}, is the traditional method of choice that has dominated the industry~\cite{featuretools,getml}. Propositionalization is essentially a data preprocessing routine where relational substructures get extracted and aggregated into standard statistical (tabular) attributes corresponding to various select-join-aggregate (SQL) routines in the database setting. 
Building on decades of practice, the resulting (statistical) models using the resulting attribute vectors typically perform very well. However, their representation learning capabilities are principally limited, as the preprocessing (denormalization) step \textit{necessarily} introduces an information loss.

\subsection{Neuro-symbolic models}

An interesting area on the intersection of proper relational (logical) representations and deep learning is known as Neural-Symbolic Integration~\cite{hammer2007perspectives}. There is a (small) number of neuro-symbolic frameworks that operate with some (subset of) FOL representation, effectively covering the relational databases while marrying the principles of neural networks through deep integration, such as Neural Theorem Provers~\cite{Rocktaschel}, Logic Tensor Networks~\cite{Serafini2016b}, or Lifted Relational Neural Networks~\cite{sourek2018lifted}. These methods are, in theory, capable of actual deep learning from relational databases. However, to the best of our knowledge, none of these methods scales to real-world database sizes due to the complexity associated with their FOL-based foundations, except for those that follow some form of the propositionalization scheme under the hood, such as~\cite{CILP}.

\subsection{Deep relational models}

The closest related work consists of extending standard neural models towards relational representations. 
The most prominent models in this category are Graph Neural Networks (GNNs)~\cite{wu2020comprehensive} designed for end-to-end learning with graph-structured data. There are currently hundreds of the original GNN model~\cite{scarselli2008graph} variants, some of which are close in spirit to our proposal, particularly some of the hyper-graph~\cite{feng2019hypergraph} and multi-relational~\cite{schlichtkrull2018modeling} extensions towards knowledge-graph applications~\cite{wang2017knowledge}. Nevertheless, the graph-based view adopted within this stream of research is generally not concerned with the salient features specific to {relational} \textit{databases}, particularly with the rich inner structure of the individual records. 

There have been only very few works that address (some of) the database-specific aspects. Particularly, the original work of~\cite{cvitkovic2020supervised} followed by an (unsuccessful) pre-training procedure in~\cite{liu2022flaky}, and the work of~\cite{bai2021atj}, which further incorporated feature engineering and random architecture search to improve its performance. 
A different line of work has been to utilize techniques from pre-training (large) language models while treating related database tuples as sentences, similarly to the tabular models~\cite{deng2022turl}, such as in~\cite{gaur2023teaching}.

In a similar spirit, the authors of~\cite{vogel2023towards} presented a (draft) vision for foundational database models, later shifting focus to scaling up GNNs for the task in~\cite{hilprecht2023spare} by leveraging symmetries. Likewise, a recent position paper of~\cite{fey2024position} aimed to establish ``relational deep learning'' as a new machine learning subfield while introducing a framework for benchmarking the GNN models,\footnote{focusing heavily on the temporal dimension of database records, which we explore experimentally in App.~\ref{app:ablations}} such as~\cite{you2024beyond},~\cite{zhang2023gfs}, and~\cite{zahradnik2023deep}.

\subsection{Our Contributions} \label{sec:constributions}
Our work can be seen as a continuation of these deep relational learning efforts, most notably the work of~\cite{zahradnik2023deep} that this paper directly expands. Particularly, we extend the existing GNN paradigm by tightly integrating the Transformer architecture into the relational message-passing scheme.
Thus, apart from proper treatment of the inter-relational structure, we also incorporate, in the spirit of the tabular Transformers, the \textit{intra-relational} structure of the attributes, embedded end-to-end within the same learning scheme. Covering the GNN efforts as a special case, we introduce the most complete framework for deep learning with \textit{actual} relational (SQL) databases, demonstrating superior results over the widest range of available benchmark datasets reported thus far.

\section{Background}
\label{sec:background}

\subsection{Relational Databases}
\label{sec:databases}

The principles of relational databases are formally based on the \textit{relational model}~\cite{codd1990relational}, rooted in FOL~\cite{gallier2015logic}, providing a unified declarative specification for managing structured data, irrespective of the particular software implementation. This abstraction allows the definition of any database as a collection of $n$-ary relations defined over the domains of their respective attributes, managed by the RDBM system to ensure consistency of the data with the integrity constraints of the logical database schema.
The key concepts to be used in this paper are as follows.

\subsubsection{Relation (Table)} Formally, an $n$-ary relation $R_{/n}$ is a subset of the Cartesian product defined over the domains $D_i$ of its $n$ \textit{attributes} $A_i$ as $R_{/n} \subseteq D_1 \times D_2 \times \dots \times D_n$, where $D_i = \mathsf{dom}(A_i)$. Each relation $R$ consists of a heading (signature) $R_{/n}$, formed by the set of its attributes, and a body, formed by the particular attribute values, which is commonly viewed as a \textit{table} $T_R$ of the relation $R$.

\subsubsection{Attribute (Column)} \textit{Attributes} $\mathcal{A_R} = \{A_1, \ldots, A_n\}$ define the terms of a relation $R_{/n}$, corresponding to the \textit{columns} of the respective table $T_R$. Each attribute is a pair of the attribute's name and a \textit{type}, constraining the domain of each attribute as $\mathsf{dom}(A_i) \subseteq \mathsf{type}(D_i)$. An attribute \textit{value} $a_i$ is then a specific valid value from the respective domain of the attribute $A_i$.
% \todo{check attribute ordering to be consistent with the Transformers PosEnc later}

\subsubsection{Tuple (Row)} An $n-$\textit{tuple} in a relation $R_{/n}$ is a tuple\footnote{the ordering is instantiated through the naming of the attributes} of attribute values ${t_i} = (a_1, a_2, \ldots, a_n)$, where $a_j$ represents the value of the attribute $A_j$ in $R$. The relation can thus be defined extensionally by the \textit{unordered} {set} of its {tuples}: $R = \{t_1, t_2,\ldots, t_m\}$, corresponding to the \textit{rows} of the table $T_R$.

\subsubsection{Integrity constraints} Besides the domain constraints $\mathsf{dom}(A_i)$, the most important integrity constraints are the primary and foreign \textit{keys}. A \textit{primary} key $PK$ of a relation $R$ is a minimal subset of its attributes $R[PK] \subseteq \mathcal{A_R}$ that uniquely identifies each tuple: 
$$\forall t_1, t_2 \in R:~ (t_1[PK] = t_2[PK]) \Rightarrow (t_1 = t_2) \,.$$ 
A \textit{foreign} key ${FK}_{R_2}$ in relation $R_1$ then refers to the primary key ${PK}$ of another relation $R_2$ as 
$$\forall t \in R_1:~ t[FK] \in \{t'[PK] \mid t' \in R_2\} \,.$$
This constitutes the inter-relations in the database, with the RDBMs managing the \textit{referential integrity} of ${T_{R_1}}[FK] \subseteq {T_{R_2}}[PK]$.

\subsection{Deep Learning}
\label{sec:deep_learning}

Deep learning~\cite{goodfellow2016deep} is a paradigm characterized by the use of \textit{gradient descent} to optimize parameters of nested functions, commonly viewed through their computation graphs, referred to as \textit{neural networks}. The main conceptual idea lies in learning \textit{latent representations} of the data corresponding to the inner layers of the networks, generally constrained to the form of fixed-size numeric tensors, which restricts directly applying deep learning to relational databases. While passing beyond that limitation, we will generalize upon concepts known from two neural architectures that address two forms or related (simpler) structured representations of \textit{sequences} and \textit{graphs}.

\subsubsection{Transformers}

The Transformer~\cite{vaswani2017attention} is a popular \textit{sequence-to-sequence} model, relying primarily on the ``{attention}'' mechanism for inter-relating the given sequence tokens $x_1, \dots, x_n$. Each input token $x_i$ here is \textit{embedded} into a continuous vector representation: ${E}(x_i) \in \mathbb{R}^d$, and combined with a ``positional encoding'' capturing its positional role: $E'(x_i) = E(x_i) + \mathsf{pos}(x_i)$.
The \textit{self-attention} mechanism then inter-relates all pairs of the input tokens to update their values as 
$$X' = \mathsf{attn}(Q, K, V) = \mathsf{softmax}\left(\frac{Q \cdot K^T}{\sqrt{d_k}}\right) V \,,$$ 
where $Q$, $K$, and $V$ are the so-called ``{\textit{query}}'', ``{\textit{key}}'', and ``{\textit{value}}'' matrix projections (``roles'') of the input embeddings $E'(X)$. This efficient matrix computation can (optionally) be further repeated in parallel with separate $Q, K, V$ projection matrices (multi-head attention).
% across multiple ``heads''.

In addition to the self-attention, Transformers employ \textit{cross-attention} for tasks involving two distinct streams of sequences. In cross-attention, the query matrix $Q$ is derived from the target ($t$) sequence {decoder}'s input, while the key $K$ and value $V$ matrices are derived from the source ($s$) sequence {encoder}'s output as $X' = \mathsf{softmax}\left(\frac{Q_t \cdot K_s^T}{\sqrt{d_k}}\right) V_s$.
In either case, the updated values $X'$ then position-wise pass through two standard feed-forward network (FNN) layers: $\mathsf{FNN}(x_i') = W_2 \cdot \mathsf{ReLU}(W_1 \cdot x_i' + b_1) + b_2$, followed by layer normalization to reduce internal covariate shift, and residual connections for improved gradient propagation.
% : $LN(x_i') = \frac{x_i' - \mu}{\sigma} \odot \gamma + \beta$, where $\mu$ and $\sigma$ are the mean and standard deviation of the input $x_i'$, and $\gamma$ and $\beta$ are learnable scale and shift parameters. Finally, residual connections~\cite{} are also used throughout the module to facilitate better gradient propagation.

\subsubsection{Graph Neural Networks}

% \todo{zminit tenhle message/update dal u combine/transform...ze na nem stavime}

GNNs are a general class of neural models aimed at \textit{graph-structured} data using the concept of (differentiable) \textit{message-passing}~\cite{wu2020comprehensive}. Given an input graph $G = (\mathcal{V}, \mathcal{E})$, with a set of nodes $\mathcal{V}$ and edges $\mathcal{E}$, let $h_v^{(l)} \in \mathbb{R}^{d^{(l)}}$ be the vector representation (embedding) of node $v$ at layer $l$.
The general concept of GNNs can then be defined through the following sequence of functions:

\begin{enumerate}
    \item[(i)] \textit{Message} function $M^{(l)}: \mathbb{R}^{d^{(l)}} \times \mathbb{R}^{d^{(l)}} \to \mathbb{R}^{d_m^{(l)}}$ computes messages for each edge $(u, v) \in E$:
    \begin{equation*}
        m_{u \to v}^{(l)} ~=~ M^{(l)}(h_u^{(l)}, h_v^{(l)}) \,.
    \end{equation*}
    
    \item[(ii)] \textit{Aggregation} function $A^{(l)}: \{\mathbb{R}^{d_m^{(l)}}\} \to \mathbb{R}^{d_m^{(l)}}$ aggregates the messages for each $v \in V$: 
    \begin{equation*}
       M_v^{(l)} ~=~ A^{(l)}\left(\{m_{u \to v}^{(l)} ~|~ (u, v) \in E\}\right) \,.
    \end{equation*}
    
    \item[(iii)] \textit{Update} function $U^{(l)}: \mathbb{R}^{d^{(l)}} \times \mathbb{R}^{d_m^{(l)}} \to \mathbb{R}^{d^{(l+1)}}$ updates representation of each $v \in V$:
    \begin{equation*}
        h_v^{(l+1)} ~=~ U^{(l)}(h_v^{(l)}, M_v^{(l)}) \,.
    \end{equation*}
\end{enumerate}

The particular choice of the message, aggregation, and update functions then varies across specific GNN models, which are commonly composed of a predefined number $L$ of such layers, enabling the message-passing to propagate information across $L$-neighborhoods within the graph(s). Note that the attention module of the Transformer follows the same schema while assuming a fully connected graph.

\section{Proposed Architecture}
\label{sec:blueprint}

In this section, we describe the proposed learning representation and the relational message-passing architecture designed for end-to-end deep learning of Transformers from databases.

\subsection{Data and Learning Representations}
\label{sec:representations}

\subsubsection{Nested hypergraphs}

In order to directly follow the inductive bias of the relational database model (Sec.~\ref{sec:databases}), we consider the learning representation of a database as a \textit{two-level multi-relational hypergraph}, where (i) each relation $R_{/n}$ forms $n$-ary hyperedges corresponding to the $n$-tuples \textit{intra-relating} its attributes $\{t_i=(a_1, a_2, \dots, a_n)\}$, and (ii) each pair $R_1,R_2$ of such relations \textit{inter-related} through the foreign key constraints ${{R_1}}[FK_{R_2}] \subseteq {{R_2}}[PK]$ forms another set of {hyperedges} from the respective tuple pairs $\{(t^1\cup{t^2}) ~|~ t^1 \in R_1, t^2 \in R_2,~t^1[FK_{R_2}] = {t^2}[PK]\}$. Note we consider all the tuple attributes $(a^1_1,\dots, a^1_n \cup a^2_1,\dots, a^2_m)$ to form the link, and not just their keys, as these may also be composite, possibly spanning the whole tuple as a corner case of $R[PK] = \mathcal{A_R}$, hence the forming of \textit{hyperedges} instead of just edges here. 

Additionally, for each such foreign-key tuple pair $(t^1, {t^2})$, we also consider the ``reverse'' hyperedge $({t^2}, t^1)$ to be able to fully propagate learning representations throughout the database, irrespective of the (ad-hoc) ordering choices of the database designer.

We then use the tuple pairs of $(t^1, {t^2})$ and $({t^2}, t^1)$ to build a bi-directional bi-partite hypergraph, connecting the tuples of the individual relations ${R_1}_{/n}, {R_2}_{/m}$, for each foreign key constraint in the database schema.

\subsubsection{Schema detection}

We aim at direct deep learning from raw database storage systems with as little preprocessing as possible while retaining the proper relational model semantics~\cite{codd1990relational}, for which we consider the relations' {attribute} values $a_i$ as the minimal processing unit, building on the formal assumption of \textit{atomicity}~\cite{codd1990relational}.
However, the current RDBMSs do not preserve the respective attribute type semantics required for deep learning. For instance, for integer-type (``int'') columns, the information on whether the data contained are of nominal, ordinal, or cyclic nature is missing. Similarly, string-type (``varchar'') columns may either contain actual text or encode discrete categories. However, such information is crucial to properly process the data with the neural models.
 
A distinction must also be made about attributes that form the key constraints as to whether they convey actual information or serve merely the referential purpose. To resolve such issues while avoiding manual data preprocessing, we have built an automated procedure that attempts to determine all such information from the database schema based on a combination of simple heuristics and selected data statistics.
% The output from the procedure is an annotated, deep learning-compatible database schema, which may optionally be (manually) amended if some additional domain knowledge for the given database is present. This is also useful for correcting databases with invalid designs, e.g. missing key specification(s).
Once the schema (Sec.~\ref{sec:databases}) is detected with all the attribute $A_i \in \mathcal{A}$ types $\mathsf{type}(D_i)$ determined, we first proceed with their \textit{encoding} to numerical values. Notably, we (optionally) transform the textual types with a pre-trained language model, particularly Sentence-BERT~\cite{reimers2019sentence} (App.~\ref{app:ablations}).

We then continue with \textit{embedding} of the attributes in an appropriate fashion. Particularly, following methods from the tabular Transformers (Sec.~\ref{sec:related}), we use a simple lookup table that stores embeddings of the detected categorical types, and ``stack'' or ``linear'' embedding of the numeric types (see App.~\ref{app:embedding} for details). Additionally, we (optionally) include the cyclic (``date/time'') types with a special embedding respecting the periodic structure of the timestamp~\cite{hu2024pytorch}. Importantly, each attribute has its own embedding function to allow for separate latent spaces.

\subsubsection{Data loading}

For machine learning, we need to establish what constitutes the learning samples $(x_i,y_i)$ in the given relational setting. In this paper, we consider the standard (self-)supervised scenario where a single attribute $A_j$ of a single target relation $R$ forms the output labels $y_i$. Nevertheless, in contrast to the (classic) tabular setting, the input examples $x_i$ can no longer be considered as i.i.d. tuples.
% , as they take the form of the multi-relational hypergraphs defined above. 

There are generally two cases: either (a) the database contains separate relational samples where each row $t_i$ of the target table $T_R$ belongs to a single learning instance $x_i$, or (b) the database cannot be split into such separate components, with $x_i$ possibly spanning the whole hypergraph structure. To extract batches of the learning samples $(x_i,y_i)$, irrespective of the structure, we follow a simple breadth-first-search (BFS) procedure, starting from each row $t_i$ of the target table $T_R$ and expanding over all the tables related through the foreign key constraints, in both the referenced and the referencing directions, while checking for loops.\footnote{Due to the possible interdependence between the samples, care must be taken to prevent information leakage about the labels, for which we mask out all target labels from the target column $A_j$ of $T_R$ when processing the samples. Overlooking this precaution led to some inappropriate accuracy reports in some of the related works.}

\subsubsection{Data sampling}
A salient feature of relational databases is that they can be very large, for which we optionally allow to run the loading natively \textit{in-database} through recursive SQL (self-)joins
% \footnote{Note that the purpose of the joins here is \textit{not} to preprocess (denormalize) the data into a single (universal) relation (as in the propositionalization), but merely to explore the primary-foreign key sub-structures.} 
with which minibatches of the hypergraph samples $\{(x_i,y_i)\}$ may be fetched into memory in a lazy fashion (with caching) from the, possibly remote, RDBMS. To make sure that the resulting hypergraph samples fit into memory, particularly in the (b) case, we (optionally) bound the BFS with a depth limit.\footnote{In inductive learning settings, this limit can be set to correspond to the perimeter of the relational receptive field of the subsequent neural message-passing, corresponding e.g. to the number of layers in GNN models (Sec.~\ref{sec:deep_learning}), without loss of information.} 

Nevertheless, in the (most) cases where the whole database simply fits into memory, the whole hypergraph structure can be conveniently loaded and accessed with the more flexible neighborhood \textit{sampling} techniques~\cite{hamilton2017inductive}. Particularly, we utilize the heterogeneous graph sampling routine introduced in~\cite{Hu2020}, which proved most suitable for our relational setting.

\subsection{Neural Architecture Space}
\label{sec:architecture}

\begin{figure*}[p]
    \centering
    \includegraphics[width=0.955\textwidth]{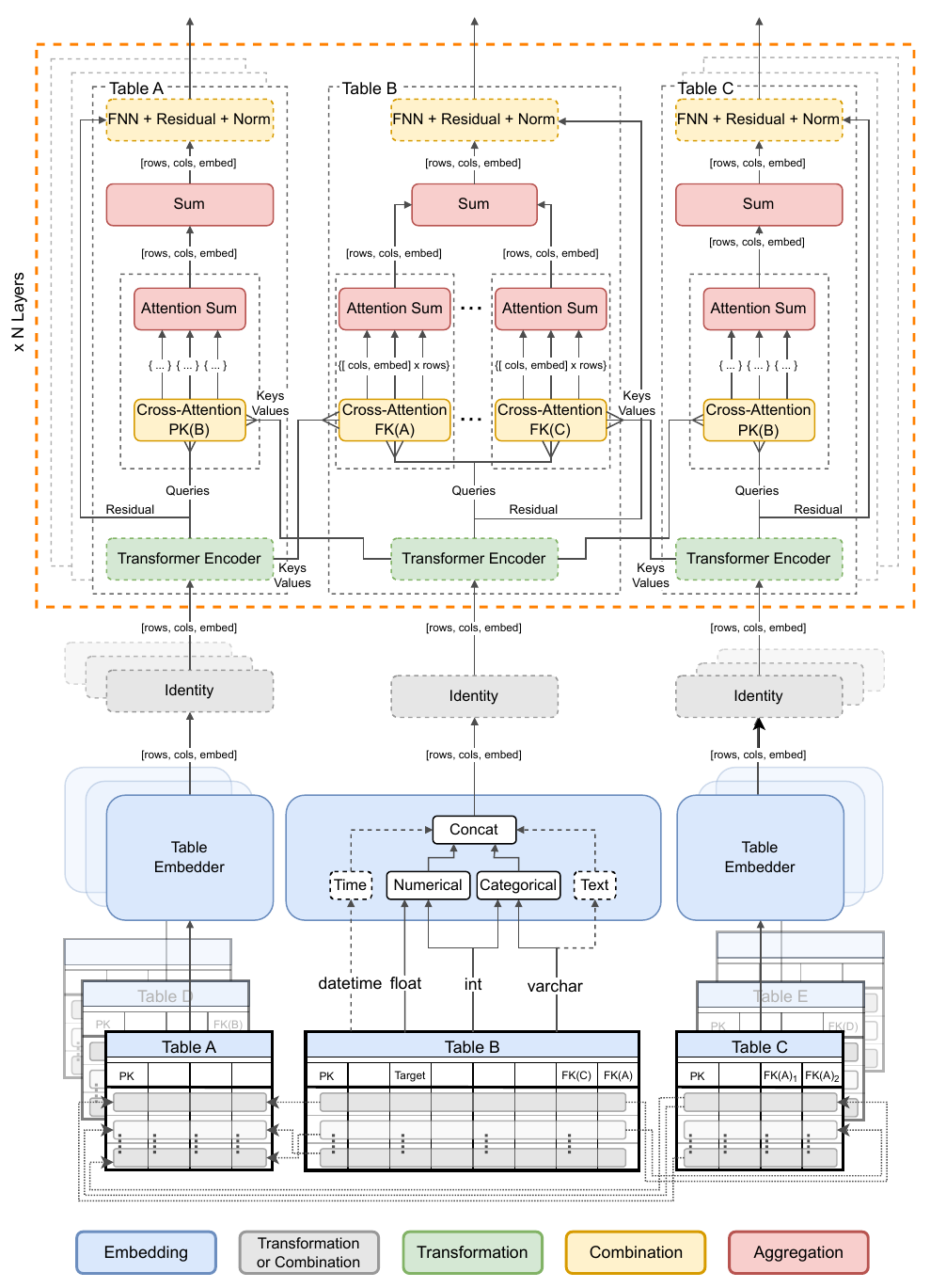}
    \caption{The relational message-passing scheme of the proposed neural architecture space, instantiated with operations of the leading \textsc{DBFormer} model.}
    \label{fig:blueprint}
\end{figure*}

To natively facilitate deep learning on the two-level hypergraph structure of the relational model (Sec.~\ref{sec:representations}), we introduce a general two-level neural message-passing scheme composed of modular differentiable parameterized operations defined on the levels of (i) individual \textit{attributes} (ii) and (sets of) related \textit{tuples}. We further divide these operations w.r.t. their input-output characteristics into three categories:
\begin{enumerate}
    \item[(i)] standard \textit{Transformations}
    $$X \overset{1 : 1}{\mapsto} Y \,,$$

    \item[(ii)] $n$-ary \textit{Combinations}
    $$(X_1, X_2, \dots, X_N) \overset{N : 1}{\mapsto} Y \,,$$

    \item[(iii)] permutation-invariant \textit{Aggregations}
    $$\{X_1, X_2, \dots, X_M\} \overset{M : 1}{\mapsto} Y \,,$$ 
\end{enumerate}
where $X$, $X_i$ and $Y$ may refer to either the attributes $a$ or the tuples $t$. 
Note that this can be seen as an extension of the ``message-aggregate-update'' paradigm of the GNNs (Sec.\ref{sec:background}). An instance of the proposed scheme is outlined in Fig.~\ref{fig:blueprint}.
% With this terminology, we may now establish the blueprint of the neural architecture space.

\subsubsection{Architecture scheme} Every instantiation of the scheme starts with the embedding (Sec.~\ref{sec:representations}) transformation (``Embedder'') of the individual relation $R_{/n}$ attribute values $E(a_1),\dots,E(a_n)$, resulting into an $n$-tuple of vectors $t_i^{(0)} \in (\mathbb{R}^{d_1},\dots,\mathbb{R}^{d_n})$ per each original tuple $t_i$ from $R_{/n}$.\footnote{The attribute embedding dimensions $d_1,\dots,d_n$ within and across the relations may generally differ, so as to accommodate the possibly varying information loads in the tables, but in this paper we set them to be the same for simplicity.} Each such tuple $t^{(0)}$ then undergoes \textit{either} (i) an attribute \textit{combination} 
$$C_a : (a_1, a_2, \dots, a_n) \overset{n : 1}{\mapsto} (a') = t^{(1)}$$ 
that merges the attribute embeddings into a joint tuple embedding $t^{(1)} \in \mathbb{R}^{d_{t^{(1)}}}$ or (ii) a tuple \textit{transformation} 
$$T_t: (a_1,\dots, a_n) \overset{1 : 1}{\mapsto} (a_1',\dots, a_n') = t^{(1)}$$ 
that keeps the attribute embeddings separate as $t^{(1)} \in (\mathbb{R}^{d_{a^{(1)}}},\dots,\mathbb{R}^{d_{a^{(1)}}})$.
In either case, the resulting tuple representation $t^{(1)}$ subsequently enters the second level of neural computation where it gets combined with all the tuples related through the second type of hyperedges (Sec.~\ref{sec:representations}). Particularly, each ${t_i^{(1)}} \in R_1^{(1)}$ undergoes a tuple \textit{combination} 
$$C_t : (t^{(1)}_i,t^{(1)}_j) \overset{2 : 1}{\mapsto} t_{i_{R_2}}^{(2)} \in \mathbb{R}^{d_{t^{(2)}}}$$ 
with each $t^{(1)}_j \in R_2^{(1)}$, where $t_i[FK_{R_2}] = t_j[PK_{R_2}]$, resulting into a set of $\{t_{i_{R_2}}^{(2)}\}$ representations for each such pair of $t_i \in R_1$ and the related $R_2$. 
Each such set of the combined representations then undergoes a tuple \textit{aggregation}
$$A_t : \{t_{i_{R_2}}^{(2)}\} \overset{m : 1}{\mapsto} t_{i_{R_2}}^{(3)} \text{~,}$$
where $m = |\{t_{i_{R_2}}^{(2)}\}|$, to obtain one $t_{i_{R_2}}^{(3)}$ representation. 
Finally, we \textit{aggregate} all such tuple representations 
$$A_{t_R} : \{t_{i_{R_k}}^{(3)}\} \overset{l : 1}{\mapsto} t_i^{(4)} \in \mathbb{R}^{d_{t^{(4)}}}$$ 
from all the $l=|\{R_k\}|$ linked relations back into a single final tuple representation $t_i^{(4)}$ for each $t_i \in R_{/n}$. 
Importantly, the same computation is performed simultaneously for \textit{each} relation $R_{/n}$ in the database, and the resulting representations may be used again as input into subsequent layers of the same computation scheme in the classic spirit of deep learning.

\subsubsection{Optional operations}
Additionally, the scheme allows for optional intermediate blocks (dashed borders in Fig.~\ref{fig:blueprint}).

First and foremost, this includes a ``post-embedding'' block that addresses the outlined division into the two options of (i) attribute combination $C_a$ and (ii) transformation $T_a$ in the first step of the scheme. Notably, combining the attributes in the (i) case 
disposes of the original column structure of $R_{/n}$, reducing the data dimensionality from $R^{n\times d}$ to $R^{d}$, and turning the remainder of the scheme into a largely standard single-level heterogeneous GNN computation~\cite{zhang2019heterognn}, as explored in some of the related works (Sec.~\ref{sec:related}). Such operation can range from a simple concatenation to \textit{Tabular Transformers} that themselves combine columns into a single row embedding, such as Trompt~\cite{chen2023trompt} or TabNet~\cite{arik2020tabnet}. Opting for the (ii) transformation then retains the original tabular structure throughout the scheme, for which we utilize operations ranging from simple positional encoding to tabular Transformer blocks retaining the columns, such as the SAINT~\cite{somepalli2021saint} and TabTransformer~\cite{hu2020tabtransformer}. 

The subsequent (optional) tuple \textit{transformation} then follows the same logic while being \textit{repeatedly} applied at the beginning of each layer of the scheme, for which the chosen model has to comply with the respective interface.
Finally, the scheme allows for a closing (optional) tuple combination, facilitating a residual connection stream in the overarching relational part.

\subsection{The \textsc{DBFormer}\label{sec:dbformer}}

Technically, any differentiable parameterized operations that satisfy the corresponding input-output interface of the transformation, combination, and aggregation operators can be used in their respective places within the scheme, some of which are presented in our experiments (Sec.~\ref{sec:experiments}).
Nevertheless, we highlight one particular instantiation that we deem to most closely integrate the essence of the original Transformer architecture~\cite{vaswani2017attention} with the relational database model (Sec.~\ref{sec:background}), which we further refer to as the \textsc{DBFormer}, depicted in Fig.~\ref{fig:blueprint}.

Firstly, the model instantiates a Transformer Encoder in place of the tuple {\textit{transformation}}, facilitating \textit{self-attention} over the relations' attributes in the standard spirit of the tabular Transformers~\cite{badaro2023transformers}, but repeated across the database and over the layers, as part of the relational scheme. Secondly, the model also uses \textit{cross-attention} in place of the tuple \textit{combination} as 
$$C_t(t_i, t_j) = \mathsf{attn}(Q=t_i, K=t_j, V=t_j) \,,$$
essentially forming a Transformer Decoder from the remaining part of the scheme per each pair of interrelated relations.

We hypothesize that the cross-attention module used in this place might be able to extract the necessary \textit{latent} relational features, as exploited with the successful propositionalization methods (Sec.~\ref{sec:related}), but in a fully end-to-end fashion through gradient descent. Based on the notable expressiveness of Transformers~\cite{lindner2023tracr}, the select-join-aggregate operations normally used to construct such relational features should be well within the hypothesis space of the resulting architecture, in which we assume the query, key, and value roles of the input tokens to correspond to the foreign-key, primary-key, and column-value roles of the individual attributes, respectively.
% This particular instantiation of the blueprint is depicted in Figure~\ref{fig:blueprint}.
The idea is that the self-attention firstly transforms the tuple attributes w.r.t. each other within the tables, the cross-attention then learns their contextual interactions with attributes from the referenced tuples, and the attention-sum finally weights all their importance w.r.t. the referencing tuples.

\section{Experiments}
\label{sec:experiments}

We test\footnote{The source code for the experiments can be found at \url{https://github.com/jakubpeleska/deep-db-learning} and the web server serving the database datasets is made publicly available at \url{https://relational.fel.cvut.cz/}
% is enclosed as supplementary material and will be, together with a web server serving the database datasets, made publicly available upon deanonymization.
} a number of instantiations of the proposed scheme against representative models from the distinct related work categories (Sec.~\ref{sec:related}) through standard supervised classification and regression tasks across a wide range of diverse relational database datasets. 

\begin{table*}[t]
\centering
\caption{Classification accuracies of the tested scheme instantiations, compared against the representative models from the related areas (Sec.~\ref{sec:related})
% % (Tabular, Relational, Propositionalization, Neuro-symbolic) 
over a range of relational database benchmarks~\cite{motl2015ctu}.}
\label{tab:results-cls}
\input{data/tables/experiments-summary-cls}
\end{table*}

\subsection{Datasets}

While RDBMs are some of the most widespread data storages, publicly available relational database benchmarks are considerably scarce. There are numerous collections of classic 
tabular~\cite{UCIRepository} and structured datasets~\cite{Berant2018Wikitables,johnson2020mimic}, including graphs~\cite{snapnets,hu2021ogb}, some of which are conceptually close to the database setting~\cite{robinson2024relbench, vogel2024wikidbs}. Nevertheless, none of these provide \textit{actual} {relational database} representations.\footnote{Instead, they present simplified CSV, JSON, or XML files that do not fully represent the RDBM setting.}
Thus, as part of this work, we have re-established the most complete resource collection in this area, originally created by~\cite{motl2015ctu}, where we currently maintain over $50$ of actual (SQL) database datasets from various domains, together with historical 
scoreboards and additional statistics.\footnote{This collection also covers most of the previous benchmarks from the domain of relational learning (Sec.~\ref{sec:related}). 
% The respective website will be made publicly available upon deanonymization.
}
% This repository also covers some of the older ILP\footnote{\url{https://www.doc.ic.ac.uk/~shm/Datasets/}} and SRL\footnote{\url{https://starling.utdallas.edu/datasets/}} dataset collections. 
% The second notable resource is the SQLshare release~\cite{jain2016sqlshare}.
% An alternative line of work is an automated creation of such datasets, e.g. from Wikipedia~\cite{vrandevcic2014wikidata}.
In this paper, we narrow these down to 19 classification (Tab.~\ref{table:cls-info}) and 16 regression (Tab.~\ref{table:reg-info}) datasets, filtering out (uninteresting) databases that are either too small or too trivial to fit.
% databases with more than two tables and more than two columns to emphasize the \textit{relational} expressiveness of the proposed approach. 
The remaining datasets are of highly diverse characteristics w.r.t. their sizes, schemas, structures, and application domains, as further detailed in App.~\ref{app:datasets}.

\subsection{Related work models}
\label{sec:related-models}

As a baseline instance of the scheme, we consider a simple \textit{tabular} FNN model~\cite{kadra2021well} operating solely on the target table, i.e., ignoring all the inter-relations. This naive strategy is useful in revealing whether the given dataset task is indeed relational in nature or not.
From the statistical \textit{relational learning} (Sec.~\ref{sec:related}), we choose the state-of-the-art RDN-boost~\cite{natarajan2012gradient}, which, following the lifting strategy, can (very roughly) be seen as a relational generalization of the popular gradient-boosted trees~\cite{friedman2001greedy}.
As the \textit{propositionalization} representative, we select the FastProp algorithm followed by XGBoost~\cite{chen2016xgboost} -- a battle-proof combination as promoted in~\cite{getml}, which leads a number of the relational dataset scoreboards~\cite{motl2015ctu}.
To cover the \textit{neuro-symbolic} area, we further emulate the popular CILP++ method~\cite{CILP} by connecting propositionalization with a FNN model in a similar fashion.
We were unable to put any of the few recent deep relational learning proposals (Sec.~\ref{sec:background}) into operation, but some of the closest GNN-based works can be viewed as conceptually close to the reduced (attribute combination) variants of the scheme (Sec.~\ref{sec:architecture}).

\subsection{Scheme instantiations}
\label{sec:scheme-models}

As the space of all the possible neural models within the proposed scheme is very large, we tested only a few selected instantiations. This means selecting some particular parameterized differentiable operations in place of the initial Embedder module, and the attribute $a$ and tuple $t$ transformations $T_{a/t}$, combinations $C_{a/t}$, and aggregations $A_{t}$ (Sec.~\ref{sec:architecture}).

\begin{table*}[t]
\centering
\caption{Regression NRMSE of the tested scheme instantiations, compared against the representative models from the related areas (Sec.~\ref{sec:related}) over a range of relational database benchmarks~\cite{motl2015ctu}.}
\label{tab:results-reg}
\input{data/tables/experiments-summary-reg}
\end{table*}

\subsubsection{\textsc{DBFormer}}
This model, already detailed in Sec.~\ref{sec:dbformer} and Fig.~\ref{fig:blueprint}, consists of $N$ layers where each can be defined as 
\begin{equation*}
    C_t^{FNN+Norm} \circ A_t^{Sum} \circ A_t^{Attn} \circ C_t^{Cross-Attn} \circ T_t^{Trans.- Encoder} \,.
\end{equation*}

With this instantiation, we further tested extending the initial baseline Embedder (Sec.~\ref{sec:representations}), transforming merely the \textit{categorical} and \textit{numerical} values to embedding vectors with the use of lookup tables and linear transformations respectively, with a number of ablations described in detail in App.~\ref{app:ablations}.

\subsubsection{DB GNN}

This model can be seen as a ``reduced'' version of the proposed scheme for its use of the attribute-combination function that flattens the columns' dimension as $C_a: (a_1, \dots, a_n) \mapsto (a_1 \dots a_n)$, where $(a_1, \dots, a_n) \in (\mathbb{R}^D, \dots, \mathbb{R}^D)$ and $(a_1 \dots a_n) \in \mathbb{R}^{n*D}$. The reduced dimensionality then allows for the use of standard graph convolution modules. Particularly, we employed the SAGE~\cite{Hamilton2018} convolution, with which the $N$ repeating layers can be described as 
\begin{equation*}
    A_t^{Sum} \circ A_t^{Sum} \circ C_t^{SAGEConv} \circ T_t^{BatchNorm + ReLU} \,.
\end{equation*} 
The model uses the baseline Embedder, and the residual combination module is skipped.

\subsubsection{DB Trompt}
This instance is designed to closely follow the tabular architecture of Trompt, as introduced in~\cite{chen2023trompt}. The Trompt Encoder is used once at the beginning as the ``post-embedding'' (Sec.~\ref{sec:architecture}) module to transform the data. The $N$ repeating layers then have a simple definition of
\begin{equation*}
A_t^{Sum} \circ A_t^{Sum} \circ C_t^{AddMean} \text{ ,}
\end{equation*}
where the tuple transformation and closing combination modules are skipped, and 
\begin{equation} \label{eq:addmean}
    C_t^{AddMean}(t_i, t_j) = t_i + \frac{1}{dim(t_j)} {\sum_{a_k \in t_j} a_k} \,.
\end{equation} 

Notably, the model utilizes the Trompt Decoder as a prediction head and has a custom Embedder that extends the baseline by following the categorical embeddings with Layer Normalization~\cite{ba2016layer}. It also uses linear transformation of numerical values followed by a ReLU activation and Layer Normalization.

\subsubsection{DB TabNet} 

Another tested instance based on a tabular Transformer is the DB extension of TabNet~\cite{arik2020tabnet}. The TabNet encoder is formed by a series of repeated Feature Transformers, each followed by the Attention Transformer.\footnote{For further description of the Feature and Attention Transformers, we refer to the original article~\cite{arik2020tabnet}.}

Similarly to the DB GNN, TabNet belongs to the ``reduced'' category. Its Embedder processes only the \textit{categorical} variables through the embeddings lookup table, and the \textit{numerical} variables are duplicated to the target dimension by the Stack Embedder (App.~\ref{app:ablations}). Its $N$ repeated layers can be defined as 
\begin{equation*}
    A_t^{Sum} \circ A_t^{Sum} \circ C_t^{AddMean} \circ T_t^{TabNet-Encoder} \,,
\end{equation*} 
where $C_t^{AddMean}$ is defined in Equation~\ref{eq:addmean}.

\subsubsection{DB SAINT}
The SAINT instance refers to the tabular model introduced in~\cite{somepalli2021saint}. The model takes a Transformer Encoder layer and extends it by a second block that uses ``Intersample Attention,'' the details of which can be found in the article~\cite{somepalli2021saint}.

The scheme's instance utilizes the ``SAINT Encoder'' layer as the tuple transformation operation in a mixture with the \textit{cross-attention} for the tuple combination. The model also uses the baseline Embedder with an extension that a ReLU activation function follows the linear transformation. The $N$ repeated layers can be defined as 
\begin{equation*}
C_t^{FF+Norm} \circ A_t^{Sum} \circ A_t^{Attn} \circ C_t^{Cross-Attn} \circ T_t^{SAINT-Encoder} \,.
\end{equation*}

\subsubsection{DB TabTransformer} 

The last experimental instance is based on the TabTransformer~\cite{Huang2020} model. The TabTransformer architecture preprocesses only the \textit{categorical} attributes, while \textit{numerical} attributes are simply passed through Layer Normalization. The \textit{categorical} columns are then passed through a Transformer Encoder block. 

Similarly to the TabNet instance, the Embedder uses lookup table embeddings for \textit{categorical} attributes and a Stack Embedder for \textit{numerical} attributes to avoid transformations of the values. The rest of the $N$ repeating layers are defined as follows
\begin{equation*} \label{eq:tabtransformer}
    A_t^{Sum} \circ A_t^{Sum} \circ C_t^{AddMean} \circ T_t^{Trans.-Encoder/LayerNorm}
\end{equation*}
with $C_t^{AddMean}$ defined in Equation~\ref{eq:addmean}.

\subsection{Parameterization} We follow a largely standard parameterization routine across all the methods. For the propositionalization-based related work, the number of relational features ranges around $200$, depending on the depth of a custom BFS procedure that we implemented to improve their default performance, and the boosting works with the optimized default of $lr=0.1$ and $100$ base estimators. For the neural methods, including the baseline tabular FNN, we follow a standard deep learning setup of tuning the embedding dimensions, learning rate, and batch size, detailed further in App.~\ref{app:hyperparams}.

\subsection{Results}
\label{sec:results}

Our classification and regression results with the models (Sec. \ref{sec:related-models}, \ref{sec:scheme-models}) are summarized in Table~\ref{tab:results-cls} and Table~\ref{tab:results-reg}, respectively. Firstly, we see that many of the datasets are simply not accessible (N/A) to the tabular models (Tabular), in cases where the target table does not contain any informative attributes. Nevertheless, in the few cases where it does, even simple tabular models (FNN) perform very well, in accordance with~\cite{kadra2021well}

The RDN-boost is a sophisticated SRL (Relational) method that does capture the relational inter-dependencies for which it, however, needs to set up ``modes,''~\cite{natarajan2012gradient} which we implemented in a rather straightforward fashion, possibly explaining its generally weaker performance. We note that we were unable to put the method into operation in the regression setting; hence, it is missing from the respective table.
More importantly, the method does not scale well to larger datasets, reported (also) with the missing values. This issue was partially shared with the other relational methods, too.
The getML (Fastprop+XGBoost) system~\cite{getml}, on the other hand, performed very well out-of-box, validating the strength of the propositionalization (Propos.) practice~\cite{propos}. Similarly, the propositionalization-based neuro-symbolic (Ne-Sy) approach of CILP++~\cite{CILP} performed very strongly, too.

Finally, instantiations of the proposed scheme generally displayed superior performances, with a small number of exceptions where the propositionalization shone. The overall best results were displayed by the proposed \textsc{DBFormer} model (Sec.~\ref{sec:dbformer}), demonstrating the strength of the close integration between the original Transformer architecture and the relational model. Nevertheless, the GNN instantiations, as well as the Tabular Transformer integrations with Trompt~\cite{somepalli2021saint} and TabNet~\cite{arik2020tabnet}, exhibited strong performances, too.

\section{Conclusions}
\label{sec:conclusion}

We introduced a general scheme that extends Transformers for deep learning from relational databases, utilizing a custom message-passing mechanism that adheres to the relational model of the common RDBMS. Our experiments with various instantiations of the scheme demonstrate its viability and superior performance as compared to commonly used methods from the associated fields of relational learning. 

To improve the performance even further, incorporating self-supervised pre-training, in the spirit of the tabular models (Sec.~\ref{sec:related}), for domain transfer across different \textit{databases} seems like a promising avenue for future work.

\appendices
\section{Experimental Setup}
\label{app:setup}

The output of the last layer, produced for the target table, is flattened if necessary and processed by a FNN prediction head with $M$ layers, with each of the hidden layers followed by ReLU activation and, optionally, Batch Normalization~\cite{ioffe2015batch}. For the standard gradient descent training in the classification tasks, the FNN output feeds into {cross-entropy} loss and MSE loss for the regression tasks, respectively.

For the metrics used in the results reporting, we simply leverage accuracy for the classification tasks and, to provide a somewhat comparable metric, a ``Normalized Root Mean Squared Error'' (NRMSE) is used across the regression tasks. The $NRMSE$ function is defined as
\begin{equation}
    NRMSE(y, \hat{y}) = \frac{RMSE(y, \hat{y})}{\Bar{y}} \,,
\end{equation}
where $\Bar{y}$ is the {mean} of all the training target values, and $RMSE$ function is defined as
\begin{equation}
    RMSE(y, \hat{y}) = \sqrt\frac{\sum_{i=1}^{n}(y_{i} - \hat{y})^{2}}{n} \,.
\end{equation}

\subsection{Environment}
\label{app:environment}

All the executed experiments discussed in Section~\ref{sec:experiments} used a simple hyperparameter optimization pipeline. The pipeline consisted of Ray~\cite{moritz2018ray}, used for the distribution of resources and model training management; Optuna~\cite{akiba2019optuna}, used for searching over the hyperparameter space; and MLFlow~\cite{zaharia2018accelerating}, used for aggregating the parameters and metrics.

As for hardware, the training runs were split into two categories based on the dataset size, more precisely based on the number of rows in the target table (App.~\ref{app:datasets}). The runs on the datasets with less than or equal to 10,000 rows were trained on a single core of the \texttt{AMD EPYC 7742 64-Core} Processor and runs on larger datasets were executed on \texttt{NVIDIA A100-SXM4 40GB} GPU with a maximum of 4 runs sharing a single GPU.

\subsection{Hyperparameters}
\label{app:hyperparams}

There were 16 runs per model and dataset executed as part of the hyperparameter search, each running for 4000+ training steps\footnote{With an exception of models that reached a hard training limit of 2 hours, however, this limit was surpassed on only the most extensive datasets such as ``tpcd'' (App.~\ref{app:datasets}) with large models. Nevertheless, extending this limit possibly allows for future improvements.} on a standard 70:30 training-validation split. All the neural models used vanilla Adam~\cite{Kingma2014} optimizer with a learning rate set as a hyperparameter on a logarithmic space within $\langle 0.00005, 0.002 \rangle$. The heterogeneous graph sampling routine (HGSampling), as described in Section~\ref{sec:architecture}), facilitated the data sampling where the batch size was parametrized by the dataset size, with a hyperparameter scale factor from an exponential space in the interval $\langle 1, 2^8 \rangle$, and limited to a value of $B$, where $B \in 2^n$ and $n \in {4, 5, \dots, 14}$; hence the batch size always remained in the interval of $\langle 16, 16384 \rangle$. The embedding dimension $D$ was also a hyperparameter in the search space, defined as a choice from the set of $\{16, 32, 64\}$. The number of layers $N$ inside the scheme's instances was set as a random integer from $\{1, 2, 3, 4, 5\}$.
The decision-making decoder {FNN} head was parametrized by the number of linear layers $M$ that was 1, 2, or 3, where each hidden layer had $64$ channels and a flag whether to use the ``Batch Normalization.''

\section{Datasets}
\label{app:datasets}

The database datasets~\cite{motl2015ctu} used for the classification and regression tasks can be viewed in Tables \ref{table:cls-info} and \ref{table:reg-info}, respectively. The tables contain statistics about the relational databases that they represent: `Num. Rels.' - number of relations inside the database, `Num. Edge. Types' - number of primary, foreign key pairs, `Num. Targ. Cols.' - number of non-key columns in the target table, `Avg. Targ. Edges' - the average number of references from a single target table row to other tables, `Total Num. Rows' - the overall number of rows in all tables of the database, such as `Total Num. Edges' - the overall number of primary, foreign key pairs between all tables of the database, `Text Col.' - whether the database contains non-key text attribute, and `Time Col.' - whether the database contains \textit{datetime} attribute.

\begin{table}[ht]
\centering
\caption{A list of \textit{classification} datasets used in the experiments with the respective statistics.}
\label{table:cls-info}
\input{data/tables/experiments-dataset-info-cls}
\end{table}

\begin{table}[ht]
\centering
\caption{A list of \textit{regression} datasets used in the experiments with the respective statistics.}
\label{table:reg-info}
\input{data/tables/experiments-dataset-info-reg}

\end{table}

\section{\textsc{DBFormer} ablation studies}
\label{app:ablations}

In this appendix section, we report the ablations performed with the main \textsc{DBFormer} model. The ablations are aimed to assess the sensitivity of the results w.r.t. (i) the selection of the initial embedding and (ii) the selection of the hyperparameters.

\subsection{Embedders}
\label{app:embedding}

The initial processing of data can often significantly influence the effectiveness of a model. Building on the work done in the field of tabular models (Sec.~\ref{sec:related}), there is a variety of possible approaches. The categorical variables are almost always encoded with a simple embedding lookup table, with the exception of the models that do not use {categorical} variables at all, e.g., Excelformer~\cite{chen2023excelformer}. Nevertheless, for the other variable types, several options may be considered.

\begin{enumerate}
    \item \textit{Stack Embedder}: the simplest option to increase the dimensionality of the \textit{numeric} attributes is to copy the value $D$ types in the embedding vector, where $D$ is the target dimension of the embeddings.

    \item \textit{Linear Embedder}: a linear layer with \textit{no activation} function, one input channel, and $D$ output channels is another common way to create the embedding vectors out of \textit{numeric} variables.

    \item \textit{Text Embeddings Transcoder}: as discussed in Section~\ref{sec:architecture}, plain text data from the database can be processed by a pre-trained language model. While it is unlikely that the language model embedding dimension will match the set-out dimension $D$, a linear layer with no activation can again be leveraged to address the dimensionality difference.
    
    \item \textit{Timestamp Embedder}: the most sophisticated embedding we considered is to account for the possible periodical information that might be encapsulated by the year, month, day, etc., of the timestamp attributes, for which the embedder first uses cyclic encoding with a combination of positional encoding to dimension $d$, where $d < D$, and only then puts the output through the linear layer to get embeddings of dimension $D$.
\end{enumerate}

The classic tabular Transformer models usually only take the opportunity to combine simple embedding for the \textit{categorical} variables with either the Stack or Linear Embedder for the \textit{numerical} variables. However, usage of the text and timestamp attributes can potentially lead to performance gains.
The \textsc{DBFormer}, representing the leading model of this paper, was thus further tested with an additional list of such embedding options as follows:

\begin{enumerate}
    \item \textit{Baseline (base)}: the embedder uses only \textit{categorical} and \textit{numerical} variables with a simple embeddings lookup table and a Linear Embedder.
    \item \textit{With Text (text)}: extends the baseline embedder with \textit{text embeddings} transformed by the Text Embeddings Transcoder.
    \item \textit{With Time (time)}: extends the baseline embedder with datetime attributes transformed by the Timestamp Embedder.
\end{enumerate}

\begin{table}[t]
    \centering
    \caption{Comparison of the baseline \textsc{DBFormer} to its version utilizing the textual embeddings. Models are only compared on the datasets containing textual non-key attributes.}
    \label{table:exp-with-text}
    \input{data/tables/experiments-text}
\end{table}

Table~\ref{table:exp-with-text} compares the performance of the baseline \textsc{DBFormer} setting to the one leveraging the textual embeddings. As can be seen, the textual embeddings significantly improve the model performance, confirming the usefulness of the information present in the often overlooked textual attributes.

\begin{table}[t]
    \centering
    \caption{Comparison of the baseline \textsc{DBFormer} to its version utilizing the timestamp embeddings. Models are only compared across the datasets containing time attributes.}
    \label{table:exp-with-time}
    \input{data/tables/experiments-time}
\end{table}

The recently proposed work of~\cite{robinson2024relbench} heavily emphasized the time dimension in the 
relational database setting. To experimentally evaluate its importance, Table~\ref{table:exp-with-time} shows the comparison of the \textsc{DBFormer} model utilizing the time attributes with the Timestamp Embedder to its baseline version. As can be seen, the Timestamp Embedder strongly improves the performance on almost all relevant datasets, again validating the importance of the information present in the time attributes. Employing both text and time attributes thus showed significant improvements in performance.

\subsection{Hyperparameter sensitivity}
\label{app:fixed-hyperparams}

All the previous experiments were carried out with the utilization of the reported hyperparameter optimization (App.~\ref{app:hyperparams}). To test the robustness of the main \textsc{DBFormer} architecture, we also present results without the hyperparameter tuning over three versions of the model listed below.
\begin{enumerate}
    \item \textsc{Large}: embedding dimension = 64, scheme $N$ layers = 4, attention heads = 4, decoder hidden layers = 2, decoder hidden channels = 64
    \item \textsc{Medium}: embedding dimension = 32, scheme $N$ layers = 3, attention heads = 4, decoder hidden layers = 2, decoder hidden channels = 64
    \item \textsc{Small}: embedding dimension = 16, scheme $N$ layers = 2, attention heads = 2, decoder hidden layers = 2, decoder hidden channels = 32
\end{enumerate}
All three models were trained with a learning rate of 0.0001 using the vanilla Adam optimizer. The dropout rate inside the attention modules was set to 0.1, and all the decoder heads utilized the Batch Normalization. The initial Embedder module did extend the baseline with both the Text Embeddings Transcoder and the Timestamp Embedder in all cases, with all the remaining settings (App.~\ref{app:setup}) being fixed.

\begin{table}[t]
    \centering
    \caption{Classification accuracies of hyperparameter optimized DBFormer* compared to the instances with fixed hyperparameters.}
    \label{table:exp-fixed-hyperparams}
    \input{data/tables/experiments-fixed-hyperparms-cls}
\end{table}

The results in Table~\ref{table:exp-fixed-hyperparams} show that the \textsc{DBFormer} model keeps displaying superior results, even without the hyperparameter tuning, and demonstrates the robustness of the architecture. Notably, the \textsc{Large} model is within 3\% of the accuracy of the optimized \textsc{DBFormer*} model (highlighted in bold) on the majority of the classification datasets. The \textsc{Medium} and \textsc{Small} models then performed adequately well, even outperforming the \textsc{DBFormer*} in a few cases where the hyperparameter optimization apparently did not find the best settings (highlighted by underlining).

\ifCLASSOPTIONcaptionsoff
  \newpage
\fi

% references section
\bibliographystyle{IEEEtran}
\bibliography{bibliography}

\end{document}

%% file: data/tables/experiments-summary-cls.tex
\begin{tabular}{l|c|c|c|c||c|c|c|c|c|c}
    Category & Tabular & Relational & Propos. & Ne-Sy & \multicolumn{6}{c}{Deep Relational (ours)} \\ \hline
\textbf{Dataset / model}   & \textbf{FNN} & \textbf{RDNboost} & \textbf{getML}    & \textbf{CILP}     & \textbf{\textsc{DBFormer}} & \textbf{GNN}     & \textbf{TabNet}  & \textbf{Trompt}   & \textbf{TabTrans.} & \textbf{SAINT} \\ \hline
Carcinogenesis         & N/A        & 59.18           & 47.96           & 69.39           & \textbf{75.51}  & 69.39          & 73.47          & 69.39           & 70.41            & 72.45        \\
CraftBeer          & 11.38       & ~0.60            & ~5.39            & 11.38           & \textbf{58.08}  & 14.97          & 14.97          & 13.17           & 13.77            & 13.17        \\
Dallas             & 49.23       & 49.23           & \textbf{86.15}  & 83.08           & 61.54           & 55.38          & 66.15          & 58.46           & 56.92            & 56.92        \\
financial          & 75.49       & N/A            & \textbf{97.06}  & 79.90           & 88.73           & 78.39          & 79.41          & 78.92           & 75.98            & 74.06        \\
Mondial            & N/A        & 39.34           & N/A            & N/A            & \textbf{100.00~} & 93.44          & 96.72          & 98.95           & 94.07            & 96.72        \\
MuskSmall          & N/A        & 40.74           & 74.07           & 81.48           & 96.30           & 96.30          & 96.30          & \textbf{100.00~} & 88.89            & 88.89        \\
mutagenesis           & 96.43       & 83.93           & 80.36           & 92.86           & 96.43           & \textbf{98.21} & \textbf{98.21} & 96.43           & 96.43            & 94.64        \\
Pima               & N/A        & 68.70           & N/A            & N/A            & \textbf{83.04}  & 80.43          & 83.48          & 80.87           & 80.00            & 81.30        \\
PremierLeague         & 59.87       & 34.21           & 61.40           & 73.68           & \textbf{99.53}  & 82.49          & 71.69          & 59.76           & 66.25            & 59.18        \\
Toxicology         & N/A        & 56.86           & 63.73           & 67.65           & \textbf{73.53}  & 70.59          & \textbf{73.53} & 71.57           & 71.57            & 71.57        \\
UW\_std            & 92.79       & 91.57           & 69.88           & 66.27           & \textbf{97.37}  & 98.06          & 86.73          & 85.98           & 86.90            & 93.39        \\
WebKP              & N/A        & N/A            & \textbf{59.70}  & 57.41           & 56.40           & 56.16          & 53.55          & 54.30           & 52.13            & 60.12        \\
DCG                & N/A        & 50.15           & 85.84           & 73.45           & 98.82           & 62.24          & 94.10          & \textbf{100.00~} & 69.91            & 64.60        \\
Same\_gen          & N/A        & 14.51           & \textbf{100.00~} & \textbf{100.00~} & \textbf{100.00~} & 89.74          & 89.57          & 88.46           & 90.57            & 93.38        \\
voc                & 78.88       & 50.02           & N/A            & N/A            & \textbf{85.16}  & 79.13          & 67.90          & 68.59           & 76.34            & 74.58        \\
PubMed             & N/A        & N/A            & \textbf{85.51}  & 84.87           & 63.38           & 55.22          & 52.48          & 61.62           & 64.07            & 61.56        \\
Accidents          & 77.40       & N/A            & N/A            & N/A            & \textbf{93.20}  & 78.70          & 77.43          & 78.22           & 78.16            & 77.75        \\
imdb\_ijs          & 64.23       & 37.19           & \textbf{94.39}  & 94.36           & 93.29           & 63.73          & 64.04          & 63.27           & 64.16            & 63.51        \\
tpcd               & 20.90       & N/A            & N/A            & N/A            & \textbf{73.35}  & 22.60          & 21.19          & 21.40           & 21.00            & 21.08        \\ \hline
\textbf{Avg. Rank} & \textbf{7.53} & \textbf{8.58}     & \textbf{5.84}     & \textbf{5.58}     & \underline{\textbf{1.95}}     & \textbf{4.37}    & \textbf{4.11}    & \textbf{4.74}     & \textbf{5.00}      & \textbf{5.11} 
\end{tabular}

%% file: data/tables/experiments-summary-reg.tex
\begin{tabular}{l|c|c|c||c|c|c|c|c|c}
    Category & Tabular & Propos. & Ne-Sy & \multicolumn{6}{c}{Deep Relational (ours)} \\ \hline
\textbf{Dataset / model}   & \textbf{FNN} & \textbf{getML}    & \textbf{CILP}     & \textbf{\textsc{DBFormer}} & \textbf{GNN}     & \textbf{TabNet}  & \textbf{Trompt}   & \textbf{TabTrans.} & \textbf{SAINT} \\ \hline

Biodegradability     & 0.1873          & 0.2061          & 0.2490           & \textbf{0.1544}       & 0.1773          & 0.1701          & 0.1654          & 0.1798             & 0.1584          \\
classicmodels        & 0.5752          & 0.6461          & 1.1939          & 0.5023                & 0.4877          & 0.4606          & \textbf{0.4048} & 0.4646             & 1.0870           \\
GOSales              & N/A            & N/A            & N/A            & 0.4179                & 0.5329          & \textbf{0.3996} & 0.5194          & 0.7880              & 0.7457          \\
northwind            & 1.1036          & 1.1588          & 1.3597          & \textbf{0.4816}       & 0.7387          & 0.8007          & 0.8784          & 0.8620              & 0.9749          \\
Triazine             & N/A            & 0.1962          & 0.1781          & 0.1354                & 0.1648          & \textbf{0.1174} & 0.1687          & 0.1752             & 0.1357          \\
Basketball\_men      & \textbf{0.2043} & 0.2283          & 0.2546          & 0.2271                & 0.2275          & 0.2798          & 0.2076          & 0.2474             & 0.2569          \\
restbase             & 0.1915          & 0.1920           & 0.1989          & 0.1771                & 0.1872          & \textbf{0.1685} & 0.1834          & 0.1847             & 0.1827          \\
AdventureWorks2014   & 0.0323          & 0.0453          & 3.2931          & \textbf{0.0113}       & 0.0635          & 2.1720           & 2.9907          & 0.3383             & 2.3792          \\
FNHK                 & 0.8262          & \textbf{0.6482} & 0.6899          & 0.7965                & 0.7974          & 0.7277          & 0.8010           & 1.0024             & 0.7494          \\
sakila               & 0.5447          & N/A            & N/A            & 0.5178                & 0.4913          & \textbf{0.4654} & 0.5525          & 0.5565             & 0.5242          \\
stats                & 0.9488          & 2.5927          & 6.4693          & \textbf{0.1410}        & 1.6549          & 0.2856          & 2.9517          & 3.0027             & 2.9768          \\
Grants               & 2.4317          & N/A            & N/A            & 3.7295                & 3.7527          & \textbf{2.4288} & 3.0689          & 2.6871             & 3.2923          \\
ConsumerExpenditures & 6.3763          & 6.2638          & 7.368           & 6.3568                & 6.3594          & \textbf{6.3380}  & 6.6393          & 6.7533             & 6.7640           \\
employee             & 0.2691          & N/A            & N/A            & \textbf{0.2644}       & 0.2645          & 0.4984          & 0.2650           & 0.2646             & 0.7050           \\
SalesDB              & N/A            & N/A            & N/A            & \textbf{0.4167}       & 0.5145          & 0.5463          & 0.5076          & 0.4424             & 0.5474          \\
Seznam               & 5.3442          & N/A            & 6.1318          & 3.6561                & 3.9379          & 4.6834          & 4.3157          & \textbf{3.4137}    & 4.0425          \\ \hline
\textbf{Avg. Rank}   & \textbf{5.5000}  & \textbf{6.3125} & \textbf{7.8125} & \underline {\textbf{2.4375}} & \textbf{4.0625} & \textbf{3.3125} & \textbf{4.5000}    & \textbf{5.0625}    & \textbf{5.4375}
\end{tabular}

%% file: data/tables/experiments-dataset-info-cls.tex
\setlength\tabcolsep{5pt}
\begin{tabular}{p{0.065\textwidth}|p{0.03\textwidth}|p{0.03\textwidth}|p{0.03\textwidth}|p{0.03\textwidth}|p{0.03\textwidth}|p{0.03\textwidth}|p{0.03\textwidth}|p{0.03\textwidth}}

\textbf{Dataset} & \textbf{Num. Rels.} & \textbf{Num. Edge Types} & \textbf{Num Targ. Cols.} & \textbf{Avg. Targ. Edges} & \textbf{Total Num. Rows} & \textbf{Total Num. Edges} & \textbf{Text Col.} & \textbf{Time Col.} \\ \hline 

\multicolumn{9}{c}{Number of rows in target table: 1 - 1000} \\ \hline

Carcinoge. & 6 & 13 & 1 & 83.21 & 28.0k & 64.1k & False & False \\ 
CraftBeer & 2 & 1 & 2 & 4.32 & 2968 & 2410 & True & False \\ 
Dallas & 3 & 2 & 13 & 2.71 & 812 & 593 & True & True \\ 
financial & 8 & 8 & 4 & 1 & 1.1M & 1.1M & True & True \\ 
Mondial & 34 & 63 & 1 & 1 & 21.4k & 43.0k & True & True \\ 
MuskSmall & 2 & 1 & 1 & 5.17 & 568 & 476 & False & False \\ 
mutagen. & 3 & 3 & 4 & 26.03 & 10.3k & 15.3k & False & False \\ 
Pima & 9 & 8 & 1 & 8 & 6912 & 6144 & False & False \\ 
Prem.Leag. & 4 & 5 & 3 & 29.29 & 11.3k & 31.8k & True & True \\ 
Toxicology & 4 & 5 & 1 & 53.26 & 49.8k & 92.5k & False & False \\
UW\_std & 4 & 4 & 4 & 1.49 & 712 & 604 & False & False \\ 
WebKP & 3 & 3 & 1 & 94.16 & 81.9k & 82.6k & False & False \\ 

\hline \multicolumn{9}{c}{Number of rows in target table: 1001 - 10 000} \\ \hline

DCG & 2 & 1 & 1 & 6.31 & 8258 & 7128 & False & False \\ 
Same\_gen & 4 & 6 & 1 & 2 & 1536 & 2978 & False & False \\ 
voc & 8 & 7 & 21 & 2.58 & 29.1k & 21.0k & True & True \\ 

\hline \multicolumn{9}{c}{Number of rows in target table: 10 001 - 100 000} \\ \hline

PubMed & 3 & 2 & 1 & 52.36 & 1.1M & 1.0M & False & False \\ 

\hline \multicolumn{9}{c}{Number of rows in target table: 100 001 - 1 000 000} \\ \hline

Accidents & 3 & 3 & 19 & 2.87 & 1.5M & 2.4M & True & True \\ 
imdb\_ijs & 7 & 6 & 2 & 4.20 & 5.6M & 8.2M & True & False \\ 
tpcd & 8 & 10 & 5 & 11 & 8.7M & 27.2M & True & True \\ \hline
\end{tabular}

%% file: data/tables/experiments-dataset-info-reg.tex
\setlength\tabcolsep{5pt}
\begin{tabular}{p{0.065\textwidth}|p{0.03\textwidth}|p{0.03\textwidth}|p{0.03\textwidth}|p{0.03\textwidth}|p{0.03\textwidth}|p{0.03\textwidth}|p{0.03\textwidth}|p{0.03\textwidth}}

\textbf{Dataset} & \textbf{Num. Rels.} & \textbf{Num. Edge Types} & \textbf{Num Targ. Cols.} & \textbf{Avg. Targ. Edges} & \textbf{Total Num. Rows} & \textbf{Total Num. Edges} & \textbf{Text Col.} & \textbf{Time Col.} \\ \hline

\multicolumn{9}{c}{Number of rows in target table: 1 - 1000} \\ \hline

Biodegrad. & 5 & 5 & 2 & 20.02 & 21.9k & 33.1k & False & False \\ 
classicmod. & 8 & 7 & 2 & 1 & 3864 & 6846 & True & True \\ 
GOSales & 5 & 4 & 1 & 39.5 & 151k & 188k & True & True \\ 
northwind & 11 & 10 & 9 & 5.6 & 3308 & 7113 & True & True \\ 
Triazine & 2 & 1 & 1 & 6 & 1302 & 1116 & False & False \\ 

\hline \multicolumn{9}{c}{Number of rows in target table: 1001 - 10 000} \\ \hline

Basketball & 9 & 9 & 59 & 23.18 & 44.8k & 62.7k & True & True \\ 
restbase & 3 & 3 & 2 & 1.99 & 19.3k & 28.4k & True & False \\ 

\hline \multicolumn{9}{c}{Number of rows in target table: 10 001 - 100 000} \\ \hline

Adv.Works & 70 & 90 & 14 & 11.26 & 760k & 1.2M & True & True \\ 
FNHK & 3 & 2 & 10 & 49.9 & 2.1M & 2.1M & True & True \\ 
sakila & 16 & 22 & 2 & 3 & 47.3k & 122k & True & True \\ 
stats & 8 & 12 & 11 & 17.44 & 1.0M & 1.6M & True & True \\ 

\hline \multicolumn{9}{c}{Number of rows in target table: 100 001 - 1 000 000} \\ \hline

Grants & 12 & 11 & 9 & 6.47 & 3.0M & 5.1M & True & False \\ 

\hline \multicolumn{9}{c}{Number of rows in target table: 1 000 001 - 10 000 000} \\ \hline

Consu.Ex. & 3 & 2 & 5 & 1 & 2.2M & 2.2M & False & False \\ 
employee & 6 & 6 & 2 & 1 & 3.9M & 4.0M & True & True \\ 
SalesDB & 4 & 3 & 1 & 3 & 6.7M & 20.1M & True & False \\ 
Seznam & 4 & 3 & 2 & 1 & 2.7M & 2.6M & False & True \\ \hline

\end{tabular}

%% file: data/tables/experiments-text.tex
\begin{tabular}{l|c|c|c}
    \multicolumn{4}{c}{\textbf{Classification}} \\
    \multicolumn{4}{c}{Model accuracy in \%} \\ \hline
    \textbf{Dataset} & \textbf{Baseline} & \textbf{With Text} & \textbf{Improvement} \\ \hline
    CraftBeer & 12.57 & 58.08 & \underline{\textbf{45.51~}} \\ 
    Dallas & 55.38 & 56.92 & \textbf{1.54} \\ 
    financial & 74.02 & 78.43 & \textbf{4.41} \\ 
    Mondial & 98.94 & 98.02 & -0.92~ \\ 
    PremierLeague & 74.79 & 90.91 & \textbf{16.12~} \\ 
    voc & 79.46 & 80.20 & \textbf{0.74} \\ 
    Accidents & 77.56 & 78.30 & \textbf{0.74}\\ 
    imdb\_ijs & 64.12 & 93.29 & \textbf{29.17~} \\ 
    tpcd & 21.26 & 73.35 & \textbf{\underline{52.09}~} \\ \hline
     \multicolumn{4}{c}{\textbf{Regression}} \\
     \multicolumn{4}{c}{Model NRMSE} \\ \hline
    \textbf{Dataset} & \textbf{Baseline} & \textbf{With Text} & \textbf{Decrease} \\ \hline
    classicmodels & 0.50 & 0.50 & 0.00 \\ 
    GOSales & 0.42 & 0.26 & \textbf{-0.16~} \\ 
    northwind & 0.48 & 0.67 & 0.19 \\ 
    Basketball & 0.23 & 0.20 & \textbf{-0.03~} \\ 
    restbase & 0.18 & 0.07 & \textbf{-0.11~} \\ 
    AdventureWorks & 0.01 & 1.61 & 1.60 \\ 
    FNHK & 0.80 & 0.81 & 0.02 \\ 
    sakila & 0.52 & 0.48 & \textbf{-0.03~} \\ 
    stats & 0.14 & 0.69 & 0.55 \\ 
    Grants & 3.73 & 4.12 & 0.39 \\ 
    employee & 0.26 & 0.26 & 0.00 \\ 
    SalesDB & 0.42 & 0.13 & \textbf{\underline{-0.28}~} \\  \hline
\end{tabular}

%% file: data/tables/experiments-time.tex
\begin{tabular}{l|c|c|c}
    \multicolumn{4}{c}{\textbf{Classification}} \\
    \multicolumn{4}{c}{Model accuracy in \%} \\ \hline
    \textbf{Dataset} & \textbf{Baseline} & \textbf{With Time} & \textbf{Improvement} \\ \hline
    Dallas & 55.38 & 61.54 & \textbf{6.16} \\ 
    financial & 74.02 & 88.73 & \textbf{14.71~} \\ 
    Mondial & 98.94 & 100.00 & \textbf{1.06} \\ 
    PremierLeague & 74.79 & 99.53 & \textbf{24.74~} \\ 
    voc & 79.46 & 85.16 & \textbf{5.70} \\ 
    Accidents & 77.56 & 79.08 & \textbf{1.52} \\ 
    tpcd & 21.26 & 21.49 & \textbf{0.23} \\ \hline

    \multicolumn{4}{c}{\textbf{Regression}} \\
    \multicolumn{4}{c}{Model NRMSE} \\ \hline
    \textbf{Dataset} & \textbf{Baseline} & \textbf{With Time} & \textbf{Decrease} \\ \hline
    classicmodels & 0.50 & 0.16 & \textbf{-0.34~} \\ 
    GOSales & 0.42 & 0.17 & \textbf{-0.24~} \\ 
    northwind & 0.48 & 0.10 & \textbf{-0.38~} \\ 
    Basketball & 0.23 & 0.17 & \textbf{-0.06~} \\ 
    AdventureWorks & 0.01 & 0.05 & 0.04 \\ 
    FNHK & 0.80 & 0.06 & \textbf{-0.74~} \\ 
    sakila & 0.52 & 0.36 & \textbf{-0.16~} \\ 
    stats & 0.14 & 0.16 & 0.02 \\ 
    employee & 0.26 & 0.25 & \textbf{-0.01~} \\ 
    Seznam & 3.66 & 4.15 & 0.49 \\ \hline
\end{tabular}

%% file: data/tables/experiments-fixed-hyperparms-cls.tex
\begin{tabular}{l|c||c|c|c}
\textbf{Dataset} & \textbf{\textsc{DBFormer*}} & \textbf{\textsc{Large}} & \textbf{\textsc{Medium}}         & \textbf{\textsc{Small}}         \\ \hline
Carcinogenesis   & 75.51                & 69.39                & 68.37                 & \textbf{73.47}       \\
CraftBeer        & 58.08                & 50.30                & 52.69                 & 44.91                \\
Dallas           & 61.54                & 56.92                & 52.31                 & 55.38                \\
financial        & 88.73                & \underline{\textbf{95.44}} & \textbf{86.76}        & \underline{\textbf{89.22}} \\
Mondial          & 100.00~               & \textbf{100.00~}      & 93.44                 & 93.44                \\
MuskSmall        & 96.30                & 85.19                & 92.59                 & \textbf{96.30}       \\
mutagenesis      & 96.43                & \textbf{96.43}       & \textbf{96.43}        & \underline{\textbf{94.64}} \\
Pima             & 83.04                & \textbf{80.43}       & \textbf{81.30}        & \textbf{80.43}       \\
PremierLeague    & 99.53                & 95.45                & \textbf{99.07}        & \textbf{99.53}       \\
Toxicology       & 73.53                & \textbf{70.59}       & 68.63                 & 69.61                \\
UW\_std          & 97.37                & \textbf{97.09}       & 88.80                 & 94.17                \\
WebKP            & 56.40                & \textbf{54.84}       & \textbf{55.25}        & \textbf{55.33}       \\
DCG              & 98.82                & \textbf{98.82}       & \textbf{\underline{100.00}~} & 92.63                \\
Same\_gen        & 100.00~               & \textbf{100.00~}      & 87.98                 & 88.40                \\
voc              & 85.16                & \textbf{81.47}       & 84.87                 & 83.76                \\
PubMed & 63.38                & \textbf{62.70}       & 57.33                 & 48.15                \\
Accidents        & 93.20                & 69.19                & 77.66                 & 79.38                \\
imdb\_ijs        & 93.29                & \underline{\textbf{93.33}} & \underline{\textbf{93.32}}  & \textbf{93.15}       \\
tpcd             & 73.35                & \textbf{70.60}       & 67.00                 & 69.17  \\ \hline             
\end{tabular}